\documentclass[11pt]{article}

\usepackage{authblk}
\usepackage{times}
\usepackage{soul}
\usepackage{url}
\usepackage[utf8]{inputenc}
\usepackage{algorithm}
\usepackage{algorithmic}
\usepackage[switch]{lineno}
\usepackage{fancyvrb}

\usepackage[
  margin=0.83in,
  headheight=12pt,
  headsep=25pt,
  includefoot,
  footskip=30pt,
]{geometry}

\usepackage{ dsfont,bbold }
\usepackage{bm}
\usepackage{caption,subcaption}
\usepackage{color}
\usepackage{booktabs} 
\usepackage{multicol,multirow}
\usepackage{fontawesome5}
\usepackage{tcolorbox}
\usepackage{dashbox}
\usepackage{enumitem}
\usepackage{graphicx}
\usepackage{verbatimbox}
\usepackage{array}  

\usepackage{pgfplots, pgfplotstable}
\pgfplotsset{compat=1.17}
\usepackage{tikz}
\usetikzlibrary{arrows,positioning,backgrounds}
\usetikzlibrary{external}
\usetikzlibrary{ calc, matrix, patterns, patterns.meta }
\usetikzlibrary{shapes,arrows, fit, tikzmark}
\usetikzlibrary{calc}
\usetikzlibrary{arrows}
\usetikzlibrary{decorations.pathreplacing}

\usepgfplotslibrary{fillbetween}
\usepgfplotslibrary{statistics}
\definecolor{msdarkblue}{RGB}{36,58,94}
\definecolor{msblue}{RGB}{0,120,215}
\definecolor{msgreen}{RGB}{16,124,16}
\definecolor{msred}{RGB}{216,59,1}
\definecolor{msgray}{HTML}{DFDFDF}
\pgfplotsset{compat=1.6}

\definecolor{msdarkblue}{RGB}{36,58,94}
\definecolor{msblue}{RGB}{0,120,215}
\definecolor{msgreen}{RGB}{16,124,16}
\definecolor{msred}{RGB}{216,59,1}
\definecolor{purple}{RGB}{128,0,128}
\definecolor{msgray}{HTML}{DFDFDF}
\definecolor{Emerald}{HTML}{00A99D}
\definecolor{RubineRed}{HTML}{ED017D}
\definecolor{pink}{HTML}{FF55A3}

\newcommand{\sys}{{\textit{LM-Sys}}}
\newcommand{\app}{{\textit{App}}}

\title{Small Language Models for Application Interactions: A Case Study}
\date{May 2024}

\author{Beibin Li, Yi Zhang, S\'ebastien Bubeck, Jeevan Pathuri, Ishai Menache}
\affil{Microsoft, Redmond}

\begin{document}

\maketitle

\begin{abstract}
We study the efficacy of Small Language Models (SLMs) in facilitating application usage through natural language interactions. Our focus here is on a particular internal application used in Microsoft for cloud supply chain fulfilment. Our experiments show that small models can outperform much larger ones in terms of both accuracy and running time, even when fine-tuned on small datasets. Alongside these results, we also highlight SLM-based system design considerations.
\end{abstract}

\section{Introduction}

Large Language Models (LLMs) are becoming pervasive in assisting humans with a wide variety of tasks, such as writing documents, presenting work, coding and health assistant. Generative LLMs are being rapidly integrated in user-facing software, for answering questions and increasing productivity through simple, language based interactions with technology. One of the key operating principles behind LLMs is exploiting their ability to generalize to unseen tasks by providing examples through the prompt itself -- an approach commonly known as in-context learning. While LLMs are being designed to support larger  prompt sizes, processing very large prompts might be expensive and incur non-negligible latencies. 

In this paper, we consider the alternative of using Small Language Models (SLMs), which are being developed nowadays and open-sourced by several companies. While SLMs may fall behind LLMs for general unseen tasks, we examine using fine-tuned versions thereof to handle a specific and \emph{fixed} set of \emph{tasks}. In particular, we consider using SLMs in order to enable natural-language interactions with a given \emph{application}; an application can be simply viewed as software that includes a set of functions (or APIs) that are invoked by the user to carry out different requirements. 

\begin{figure*}
    \centering

\includegraphics[width=0.9\textwidth]{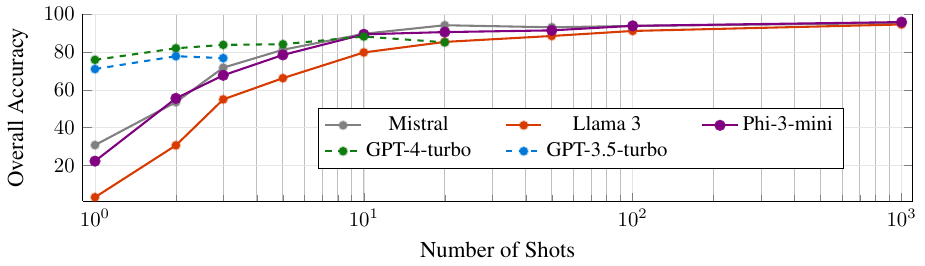}

\caption{
Overall accuracy as a function of the number of the per-task examples. For LLMs, the examples are given in the prompt, and for SLMs they are used for the offline fine tuning.}
    \label{fig:converge}

\end{figure*}

Organizations deploy customized applications in a variety of sectors, including healthcare \cite{lee2023benefits}, supply chains \cite{simchi1999designing}, and agriculture \cite{balaguer2024rag}. Our focus here on a particular application that has been developed in Microsoft for managing the server fulfillment process for Azure's supply chain -- namely, the process of sending hardware from warehouses to the data centers. The APIs of this application include adding business and operational constraints for fulfilment, generating a fulfillment plan and extracting details and insights about the plan. We build a language model system on top of this application to enable users to accomplish essential tasks, each of which utilizing one or more APIs. For example, users may interact with our system in plain English to generate a new plan, or answer what-if queries (e.g., ``what are the implications of disabling a certain warehouse"?). In this paper, we experiment with different language model technologies that can be used in our system. Our experiments show that SLMs such as Phi-3, Llama 3 and Mistral v0.2 can achieve higher degrees of accuracy than state-of-the-art LLMs, while providing outputs significantly faster. Similar observations regarding the potential benefits of SLMs have been recently reported in \cite{zhao2024lora} for a variety of benchmarks. Importantly, our paper shows that SLMs may reach these superior capabilities with a modest size of training data (see Figure \ref{fig:converge}). 


Our results, while corresponding to only a single case study, may point to an attractive option of using SLMs for interacting with applications. Aside from the potential performance benefits, the SLMs can be hosted locally on a single machine. Such system architectures can be particularly appealing for edge computations (in warehouses, farms, vehicles, etc.) where data transfers and Internet connectivity might be a bottleneck.

\section{Background and Motivation}
\subsection{Small Language Models}
Concurrent with the advent of the Transformer architecture \cite{Vas17}, a concept known as ``scaling laws" began to take shape \cite{kaplan2020scaling}. These laws suggest that as we increase computational resources or the size of neural networks, we can expect a corresponding improvement in performance \cite{hestness2017deep}. This principle has guided the expansion LLMs, with notable examples including GPT-3 and GPT-4 \cite{gpt3, gpt4}. Further refinements to these scaling laws have led to significant leaps in model capabilities \cite{hoffmann2022an}.

Recently, some research has pivoted towards another dimension of improvement: the enhancement of \emph{data quality}, which reshapes the scaling law and ultimately led us to a series of Small Language Models \cite{gunasekar2023textbooks, li2023textbooks, javaheripi2023phi, liu2023tinygsm, abdin2024phi}. These small models operate with orders of magnitude less parameters but are capable of rivaling their larger counterparts on key performance benchmarks. It's a well-established fact that better data yields better outcomes, as from dataset development \cite{raffel2020exploring}. High-quality data also results in smaller datasets \cite{longpre2023pretrainer, yu2023selective} and the opportunity for more iterations over the data \cite{muennighoff2023scaling}.

Due to their efficiency and adaptability, SLMs have emerged as an important alternative to LLMs in many resource-criticial industrial scenarios. Their surprisingly low parameter count makes them ideal for applications requiring 1) promptly response, 2) little computation overhead, 3) on-device processing of private data and customized requests.

\paragraph{The Phi models:}
One series of powerful SLMs is the ``Phi" models, which operates under the principle that ``textbooks are all you need" \cite{gunasekar2023textbooks}. These models leverage a highly curated dataset to train an efficient yet capable model. In particular, Phi-3, the latest member of the Phi series~\cite{abdin2024phi}, achieves remarkable performance with a small amount of parameters (ranging from 3.8B to 14B).

\subsection{Cloud supply chain fulfillment}
In this section, we provide some general background on the application for which we built natural language support. We then describe the interaction mode through tasks and provide examples. 

\begin{table}[]
    \centering
    \caption{Example task types and code for them.}
\begin{tabular}{l|p{4cm}|p{10cm}}
\toprule
Category & Task & Code Snippet (w/ APIs)  \\
\midrule

\multirow{2}{*}{Data Extraction}  & What is the standard deviation of supplier S's inventory in the last T weeks? & 
\begin{minipage}[t]{10cm}
\begin{verbatim}
ans = retrieve("""SELECT STDDEV(quantity)
FROM inventory WHERE supplier_id = 'S'
AND record_date >=
    NOW() -  INTERVAL T WEEK;""")
logger.log(f"The std is {ans}")
\end{verbatim}
\end{minipage}
\\
\cline{2-3}
 & What was the fraction of cross-geo shipments in the last T weeks &
\begin{minipage}[t]{10cm}
\begin{verbatim}
total = retrieve(f"""SELECT SUM(quantity) 
    FROM shipment WHERE date >= 
    NOW() - INTERVAL '{T} weeks'""")
cross = retrieve(f"""SELECT SUM(quantity) 
    FROM shipment WHERE date >= 
    NOW() - INTERVAL '{T} weeks'
    AND src_geo != dest_geo;""")
if total:
    logger.log(cross / total * 100)
else:
    logger.log("No shipments at all")
\end{verbatim}
\end{minipage}
 \\
\midrule
 
\multirow{2}{*}{Plan Generation}  & Optimize plan &  \begin{minipage}[t]{10cm}
\begin{verbatim}
model.optimize()
\end{verbatim}
\end{minipage} \\
\cline{2-3}
& Update (Execute) plan & 
\begin{minipage}[t]{10cm}
\begin{verbatim}
plan.update()
\end{verbatim}
\end{minipage}
\\
\cline{2-3}
&  Dock demand D on its ideal dock date! & 
\begin{minipage}[t]{10cm}
\begin{verbatim}
ideal_date = retrieve("""SELECT idd FROM 
    demand WHERE id = 'D';""")
demand.add_constraint(
    demand_id="D", date=ideal_date,
    enforce="Exact Match")
model.optimize()
if model.feasible:
    plan.update()
    logger.log("Plan updated with cost", 
    model.objVal)
else:
    logger.log("Sorry, impossible to "
    "dock demand D at its ideal date.")
\end{verbatim}
\end{minipage} \\
\midrule
\multirow{2}{*}{What-if Analysis}  & What are the cost implications of not using suppliers from region R on Month M? & 
\begin{minipage}[t]{10cm}
\begin{verbatim}
supplies = retrieve("""SELECT id FROM 
    supplier WHERE region = 'R';""")
if len(supplies):
    [ supply.add_constraint(
         supply_id=id, demand="*",
         date="2024-M-*", enforce="Prohibit")
       for id in supplies ]
    model.optimize()
    logger.log(f"Cost will be {model.objVal}")
    model.reset()
else:
    logger.log("No supplies in R.")
\end{verbatim}
\end{minipage}
\\
\bottomrule
\end{tabular}

    \label{tab:tasks}
\end{table}

\noindent \textbf{The application.} We consider the fulfillment management application which has been designed internally at Microsoft for managing cloud supply chain fulfillment, a critical operational phase in cloud resource management in light of the constant growth of the cloud (see \cite{optiguide} and references therein); we henceforth refer to that application simply as \app{}, as our design methodology is general and can be applied to other applications. At a high level, \app{} manages the decision making process for satisfying the demand for cloud hardware. A demand request comes from internal Microsoft organizations, such as Microsoft 365 or Azure, and it is specified in racks of server units, alongside a desired dock (or deployment) date. \app{} is in charge of generating a \emph{fulfilment plan} for a configurable time-horizon (typically in the order of weeks). For each demand request, the plan includes (i) \emph{supplier.} The hardware supplier (including warehouse location) that will be used to fulfill the demand, (ii) \emph{Shipping.} The shipping method that will be used to transfer the racks of hardware from the warehouse, and the target datacenter that will host these racks, and (iii) \emph{Scheduling.} The timing of the shipment; note that the a combination of shipping and scheduling determines the dock date. \app{} uses an optimization algorithm to automatically generate a plan while accounting for different business and operational constraints \cite{optiguide}.

\noindent \textbf{Motivation behind language models.} \app{} is consumed by \emph{planners} who periodically generate plans and oversee that the execution of the decisions is completed as planned. Planners may be interested in obtaining insights from historical plan data, understanding certain decisions, as well asking what-if questions. What-if questions may correspond to understanding the potential impact on the plan if certain conditions are changed (e.g., a supplier becomes unavailable, a certain shipping date must be enforced, etc.). Before the LLM disruption, planners would interact with engineering teams for understanding issues and answering what-if questions \cite{optiguide}. As we elaborate below, the role of language models is to facilitate the planners' job by enabling direct and efficient interaction with \app{}.

\noindent \textbf{Tasks.} 
\app{} may be used in order to carry out different \emph{tasks}. The task types can be divided into three categories:  data extraction, plan generation, and what-if analysis. Each task is carried out through one or more APIs (\app{} has over forty different APIs). We provide some task examples below.

\noindent \textbf{Data extraction:}
\begin{itemize}
    \item {[supplier]} What is the standard deviation of supplier S's inventory in the last T weeks?
    \item {[shipping]} What was the fraction of cross-geographical shipments in the last T weeks?
    \item {[scheduling]} Will demand D be deployed in its ideal dock date?
\end{itemize}

\noindent \textbf{Plan generation:}
\begin{itemize}
    \item Optimize plan while taking into account a set of new constraints C.
\end{itemize}

\noindent \textbf{What-if analysis:}
\begin{itemize}
    \item {[supplier]} What are the cost implications of not using suppliers from region R on Month M?
    \item {[shipping]} By how much would a plan cost increase if we force priority shipping for demand D?
    \item {[scheduling]} Is it possible to dock demand D in its ideal dock-date?
\end{itemize}

\noindent \textbf{Language-model system overview.} Our system uses a language model to translate a human query to the right task and produce the relevant code for executing it. We henceforth refer to the system as \sys{}. Our implementation currently supports around fifty different task types. For any supported task type, \sys{} will translate a corresponding query (given in English) to a Python snippet, which can use multiple \app{} APIs. We provide below a few examples, see also  Table \ref{tab:tasks}. 
For a new plan generation, a user may first ask the system to generate a plan while taking into account a new set of constraints. For example, the task ``Generate a plan that does not use any suppliers from region R on Month M", will first invoke an API to find the IDs of the suppliers from region R demand, and then use these IDs to add constraints prohibiting using the corresponding suppliers in the plan; the last step is generating the plan, which invokes an algorithm that will generate an optimal plan. What-if tasks produce a quantitative analysis of certain conditions are modified. Consequently, their implementation can be very similar to that of plan generation. For example, the task ``What would be the cost increase if we do not use any supply from region R in Month M", will use the same code sequence as in the previous example, except that instead of updating the plan, the system will only invoke the algorithm in order to examine the total cost and compare it with that of the current plan.

\begin{figure*}[h]
    \centering
    \includegraphics[width=0.9\textwidth]{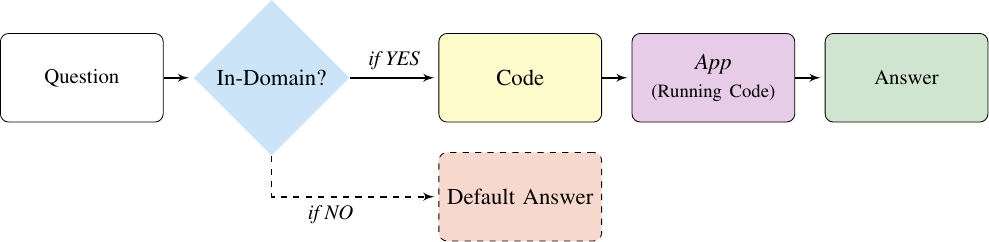}
\caption{\textbf{\sys{} logical flow}. User queries are first processed to determine if they are in-domain. For in-domain queries, \sys{} generates the relevant code snippet that is executed to provide the answer. Otherwise, the SLM will return a default response (e.g., guide the user about supported tasks.}
\label{fig:system}
\end{figure*}

\section{Language-model system design}
\begin{figure}
    \centering
    \includegraphics[width=0.99\textwidth]{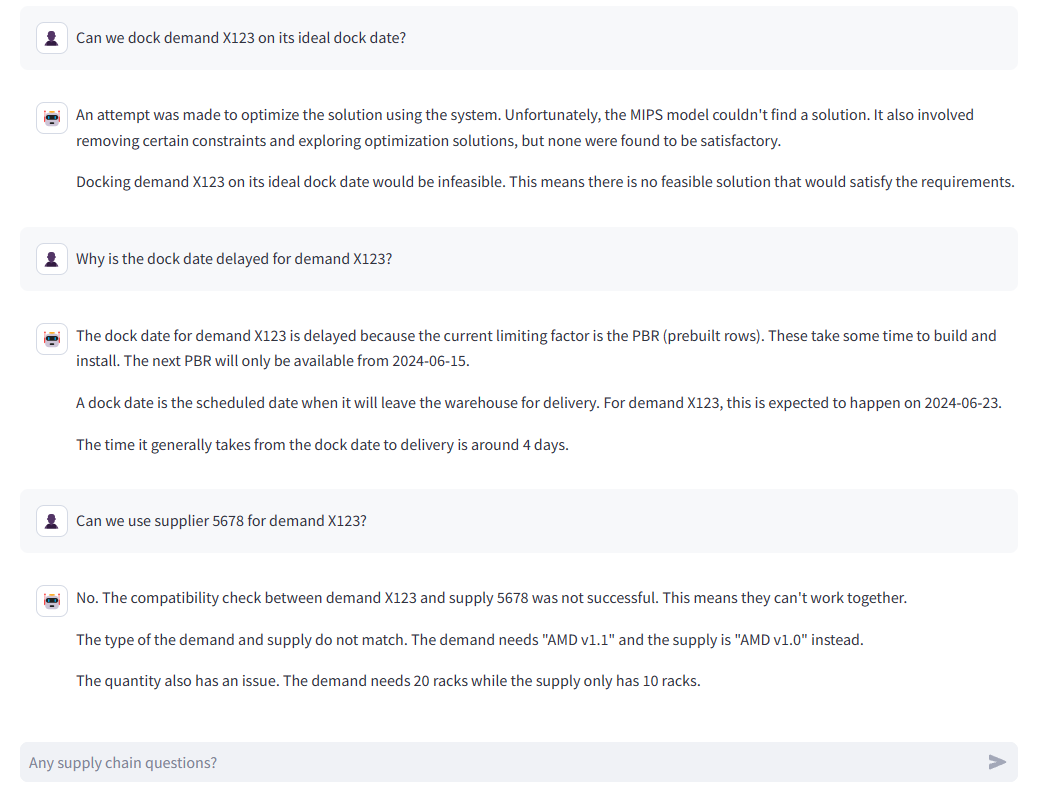}
    \caption{An example of an interaction log from production.} 
    \label{fig:screenshots}
\end{figure}

In this section we highlight some design details of \sys{}. Chronologically, our first implementation used LLMs, and only very recently we started incorporating SLMs in production. Figure \ref{fig:system} depicts \sys{} data flow. 
A screenshot of an interaction log from production is given in Figure \ref{fig:screenshots}.

\paragraph{Natural-language interaction challenges.}
Building on our production experience, we observe that users occasionally pose questions that are irrelevant to the core functionality of the application, such as requests to ``help me rewrite this email", ``how is the weather today", or simply ``how are you". One of our design goals has been 
to systematically filter out such questions. Another challenge arises from the variety of ways that a certain query can be phrased. For example, the following query, ``why is my demand not fulfilled?", ``there is no docking for my demand", and ``can you dock my demand?" all point to the same task of understanding the reasoning for an unfulfilled demand (e.g., is it possible at all to fulfill the demand, or alternatively, quantify the cost implication of forcing that demand to be fulfilled). As the number of different supported tasks increases, it becomes more difficult to include the different question variations in the prompt itself. Fine-tuning thus becomes an attractive alternative, as one can increase the number of training examples to capture the diversity.

\paragraph{SLM fine-tuning.}
In our SLM-based design, we include the entire task handling flow in a single training example, similar to the ``chain-of-thought" concept \cite{wei2022chain}. The process of generating training examples adheres to a conventional data generation pipeline. It incorporates techniques such as using example templates \cite{optiguide}, ``textbooks are all you need" \cite{gunasekar2023textbooks,li2023textbooks,abdin2024phi}, and prompt evolution from WizardLM \cite{xu2023wizardlm,luo2023wizardcoder}. To enhance the diversity and complexity of the data, we employed GPT-4 to rewrite human-posed questions, incorporating alternative writing styles including typos and distractions. 

To ensure robust results, the SLM model training employed consistent hyperparameters across all setups. We utilized an effective batch size of 16 and employed the AdamW optimizer with a stable learning rate of 0.0002. Each model was constrained to a maximum of 1024 input tokens and 500 output tokens to meet our application needs. LoRA was integrated into the training process to effectively handle the relatively small datasets available \cite{hu2022lora}. The overall training was capped at 100,000 steps of gradient updates, so that each model has the same conditions for optimizing its weights no matter how many unique training examples it has seen.  

\paragraph{Output example.} A screenshot of an interaction log from production is given in Figure \ref{fig:screenshots}.

\begin{table}[h] 
\centering 
\begin{tabular}{lc|c|cc|c} 
\toprule 
\textbf{Model} & \textbf{Method} & \textbf{Overall Acc. (\%)} & \textbf{Coder Acc. (\%)} & \textbf{F-1 (\%)} &  Cost (¢) \\ 
\midrule 
Phi-2 (2.7B) \cite{li2023textbooks} &   \multirow{6}{*}{\rotatebox{90}{\small Tuned w/ 1000-shot}} & 89.48 $\pm$ 1.46 & 91.12 $\pm$ 0.66 & 64.28 $\pm$ 35.71 & 0.18   \\ 
Phi-3 mini (3.8B) \cite{abdin2024phi} & & \textbf{95.86} $\pm$ 0.33  & \textbf{95.64} $\pm$ 0.34 & \textbf{100}  &  0.19  \\ 
Gemma (7B) \cite{team2024gemma} &  & 92.62 $\pm$ 4.06 & 92.23 $\pm$ 4.26 & 88.73 $\pm$ 25.18  & 0.25  \\ 
Mistral v0.2 (7B) \cite{jiang2023mistral} &  & 95.47 $\pm$ 0.26 & 95.23 $\pm$ 0.27 & \textbf{100} & 0.29   \\ 
Gorilla OpenFunc v2 (7B) \cite{patil2023gorilla} &  & 93.83 $\pm$ 0.94 & 94.00 $\pm$ 0.92 & 73.40 $\pm$ 26.62 &  0.21  \\ 
Llama 3 (8B) \cite{llama3} &  & 94.59 $\pm$ 0.71 & 94.85 $\pm$ 0.75 & 90.0 $\pm$ 15.66 & 0.28  \\ 
\midrule
\midrule
GPT-3.5-turbo &  \multirow{2}{*}{1-Shot}  & 71.01 & 75.21 & 80.00  &  0.23  \\
GPT-4-turbo  &  & 75.88 & 74.60 & 75.00  &    4.52 \\
\midrule
GPT-3.5-turbo  &  \multirow{2}{*}{2-Shot}  & 77.81 & 76.63 & 80.00   & 0.44 \\
GPT-4-turbo   &  & 82.00 & 81.05  &  85.71   & 8.72 \\
\midrule
GPT-3.5-turbo  &  \multirow{2}{*}{3-Shot}  & 76.78 & 75.56 &  80.00     & 0.66 \\
GPT-4-turbo  &  & 83.84 &  82.96 &  80.00    &  12.92 \\
\midrule
GPT-3.5-turbo  &  5-shot &  \multicolumn{4}{c}{Out of Memory}  \\
GPT-4-turbo  &  5-shot & 84.15 &  83.31 &  85.71    & 21.32 \\
GPT-4-turbo &  10-shot & 88.22 &  87.60  &  92.31    &  42.32  \\
GPT-4-turbo &  15-shot & 87.99 &  87.37  &  92.31    & 63.32  \\
GPT-4-turbo  &  20-shot & 85.17 & 84.39 &  92.31    &  84.32  \\
\bottomrule 
\end{tabular} 
\caption{
Accuracy and cost per query of the different models. The SLMs results include average + standard deviation over ten runs.
} 
\label{tab:rst} 
\end{table}

\section{Evaluation}

In this section we compare the different language models that we considered. They include different versions of GPT with in-context learning, and multiple fine-tuned SLMs.

\subsection{Accuracy}
\paragraph{Metrics.} Table \ref{tab:rst} summarizes the accuracy of the different models. We consider three metrics for accuracy. Analyzing production data, we observe that a non-negligible subset of queries (3-5\%) is ``out-of-domain", namely does not correspond to any of the tasks. As described above, the language model should identify such tasks and provide a `generic' answer. The \textit{F-1 Score} represents the success ratio of the identification. \textit{Coder accuracy} measures the precision of the code generation for in-domain queries, namely queries that capture supported tasks. Finally, \textit{Overall accuracy} is the ratio of queries (both out-of-domain and in-domain) for which the system produces the right output.  

\paragraph{Results.}
The results depicted in Table \ref{tab:rst} correspond to using one thousand training examples per task for the SLMs fine-tuning. For the LLMs, we report results with a varying number of in-context learning examples. At some point, the prompt size becomes a bottleneck and we cannot practically increase it further. The results indicate that Phi-3 mini and Mistral obtain the best results. The different LLM models fall short of the best SLMs.

We next zoom in on the convergence rate of the different models as a function of the number of training examples. For the GPT models, the examples are given in the prompt, and for the SLMs they are used for offline fine tuning. Figure \ref{fig:converge} shows the results, where the x-axis represents the number of shots (or training examples per task); the total number of examples is obtained by multiplying that number by the number of tasks. The results show that GPT-4 is superior until 10 shots, but then plateaus. Phi-3 and Mistral obtain high accuracy with relatively few shots, and then improve gradually with more examples.

\subsection{Performance and cost}

\paragraph{Token count and latency.}
For gradient-free LLM prompting, input token counts vary based on the number of shots. For example, for GPT-4-turbo there are approximately 4300 for 1-shot, 8500 for 2-shot, and 12400 for 3-shot. There are, on average, 72.24 output tokens with a standard deviation of 22.25.  
In comparison, the average input token count for Phi-3 is only 18.89 (std=12.88), and the output size is 108.76 (std=30.76) tokens. Due to the significantly lower average input token count, the fine-tuned Phi-3 can handle queries quickly, typically taking only a few seconds, compared to 1-2 minutes for GPT-4-turbo.

\paragraph{Inference cost.}
The cost of GPT models is calculated based on the average number of input and output tokens processed\footnote{As of May 2024, GPT-3.5-turbo-0125 costs \$0.50 per million input tokens and \$1.50 per million output tokens, while GPT-4-turbo is priced at \$10.00 per million input tokens and \$30.00 per million output tokens.}. The cost per query for an SLM is estimated by the renting price of a GPU VM per hour divided by the number of queries that can be processed per hour; we note that this cost model assumes high GPU utilization, which can be achieved, for example, by utilizing Multi-LoRA inference servers  \cite{zhao2024lora}. In our calculation, we assume a batch size of one for the SLMs (costs may be lower if we apply mini-batching). 

\paragraph{Training cost.}
Finally, the estimated training cost for SLMs is in the order of \$10 for 1000 shot training, which adds a negligible amount to the per-query cost, even when retraining is done every week.

\section{Related Work}

A preliminary version of using LLMs for interacting with the cloud supply chain fulfillment application was reported in \cite{optiguide}; that version supported a much smaller number of tasks. The \textit{Gorilla} benchmark \cite{patil2023gorilla} is used to examine language model capabilities in translating a human query to the correct function call. The setting of that benchmark is different than ours, as the set of possible function calls is not known in advance and cannot be used in the training phase. 

\bibliographystyle{unsrt}
\bibliography{ref}
\end{document}